\PassOptionsToPackage{unicode}{hyperref}
\PassOptionsToPackage{hyphens}{url}
\PassOptionsToPackage{dvipsnames,svgnames,x11names}{xcolor}
\documentclass[
  12pt]{article}

\usepackage{amsmath,amssymb}
\usepackage{iftex}
\ifPDFTeX
  \usepackage[T1]{fontenc}
  \usepackage[utf8]{inputenc}
  \usepackage{textcomp} % provide euro and other symbols
\else % if luatex or xetex
  \usepackage{unicode-math}
  \defaultfontfeatures{Scale=MatchLowercase}
  \defaultfontfeatures[\rmfamily]{Ligatures=TeX,Scale=1}
\fi
\usepackage{lmodern}
\ifPDFTeX\else  
\fi
\IfFileExists{upquote.sty}{\usepackage{upquote}}{}
\IfFileExists{microtype.sty}{% use microtype if available
  \usepackage[]{microtype}
  \UseMicrotypeSet[protrusion]{basicmath} % disable protrusion for tt fonts
}{}
\makeatletter
\@ifundefined{KOMAClassName}{% if non-KOMA class
  \IfFileExists{parskip.sty}{%
    \usepackage{parskip}
  }{% else
    \setlength{\parindent}{0pt}
    \setlength{\parskip}{6pt plus 2pt minus 1pt}}
}{% if KOMA class
  \KOMAoptions{parskip=half}}
\makeatother
\usepackage{xcolor}
\makeatletter
\ifx\textbf\undefined\else
  \let\oldparagraph\textbf
  \renewcommand{\textbf}{
    \@ifstar
      \xxxParagraphStar
      \xxxParagraphNoStar
  }
  \newcommand{\xxxParagraphStar}[1]{\oldparagraph*{#1}\mbox{}}
  \newcommand{\xxxParagraphNoStar}[1]{\oldparagraph{#1}\mbox{}}
\fi
\ifx\textbf\undefined\else
  \let\oldsubparagraph\textbf
  \renewcommand{\textbf}{
    \@ifstar
      \xxxSubParagraphStar
      \xxxSubParagraphNoStar
  }
  \newcommand{\xxxSubParagraphStar}[1]{\oldsubparagraph*{#1}\mbox{}}
  \newcommand{\xxxSubParagraphNoStar}[1]{\oldsubparagraph{#1}\mbox{}}
\fi
\makeatother

\usepackage{longtable,booktabs,array}
\usepackage{calc} % for calculating minipage widths
\usepackage{etoolbox}
\makeatletter
\patchcmd\longtable{\par}{\if@noskipsec\mbox{}\fi\par}{}{}
\makeatother
\IfFileExists{footnotehyper.sty}{\usepackage{footnotehyper}}{\usepackage{footnote}}
\makesavenoteenv{longtable}
\usepackage{graphicx}
\makeatletter
\def\maxwidth{\ifdim\Gin@nat@width>\linewidth\linewidth\else\Gin@nat@width\fi}
\def\maxheight{\ifdim\Gin@nat@height>\textheight\textheight\else\Gin@nat@height\fi}
\makeatother
\setkeys{Gin}{width=\maxwidth,height=\maxheight,keepaspectratio}
\makeatletter
\def\fps@figure{htbpp}
\makeatother

\makeatletter
\@ifpackageloaded{caption}{}{\usepackage{caption}}
\AtBeginDocument{%
\ifdefined\contentsname
  \renewcommand*\contentsname{Table of contents}
\else
  \newcommand\contentsname{Table of contents}
\fi
\ifdefined\listfigurename
  \renewcommand*\listfigurename{List of Figures}
\else
  \newcommand\listfigurename{List of Figures}
\fi
\ifdefined\listtablename
  \renewcommand*\listtablename{List of Tables}
\else
  \newcommand\listtablename{List of Tables}
\fi
\ifdefined\figurename
  \renewcommand*\figurename{Figure}
\else
  \newcommand\figurename{Figure}
\fi
\ifdefined\tablename
  \renewcommand*\tablename{Table}
\else
  \newcommand\tablename{Table}
\fi
}
\@ifpackageloaded{float}{}{\usepackage{float}}
\floatstyle{ruled}
\@ifundefined{c@chapter}{\newfloat{codelisting}{h}{lop}}{\newfloat{codelisting}{h}{lop}[chapter]}
\floatname{codelisting}{Listing}

\makeatother
\makeatletter
\@ifpackageloaded{caption}{}{\usepackage{caption}}
\@ifpackageloaded{subcaption}{}{\usepackage{subcaption}}
\makeatother

\ifLuaTeX
  \usepackage{selnolig}  % disable illegal ligatures
\fi
\usepackage[]{natbib}
\usepackage{bookmark}

\IfFileExists{xurl.sty}{\usepackage{xurl}}{} % add URL line breaks if available
\hypersetup{
  colorlinks=true,
  linkcolor={blue},
  filecolor={Maroon},
  citecolor={Blue},
  urlcolor={Blue},
  pdfcreator={LaTeX via pandoc}}

\usepackage{amsmath, amssymb, amsthm}
\usepackage{graphicx}
\usepackage{enumerate}
\usepackage{enumitem}
\usepackage{natbib}
\usepackage{url}
\usepackage{geometry}
\usepackage{booktabs} 
\usepackage{multirow} 
\usepackage{array} 
\usepackage{xcolor} 
\usepackage{colortbl}
\usepackage{makecell}
\usepackage{setspace}
\usepackage{hyperref}
\usepackage{tabularx}
\usepackage[title,toc,titletoc]{appendix}
\usepackage{bbm}
\usepackage{bm} % for bold math symbols
\usepackage{algorithm}
\usepackage{booktabs}
\usepackage{longtable}
\usepackage{siunitx}
\usepackage{multirow}
\usepackage{algpseudocode}
\usepackage{multibib}
\usepackage{etoc}
\usepackage{pifont}
\makeatletter
\newcommand{\appendixtableofcontents}{%
  \section*{Table of Contents}
  \@starttoc{apx}
}
\makeatother
\newcites{supp}{References of the supplementary material}
\usepackage{subcaption}
\newcommand{\anon}{1}

\newtheorem{theorem}{Theorem}
\newtheorem{assumption}{Assumption}
\newtheorem{remark}{Remark}
\newtheorem{proposition}{Proposition}
\newtheorem{lemma}{Lemma}
\newtheorem{corollary}{Corollary}[theorem]

\usepackage{booktabs}
\usepackage{listings}
\usepackage{fvextra}

\usepackage{fancyvrb}
\usepackage{xcolor}
\definecolor{promptbg}{HTML}{F7F7F9}
\definecolor{promptframe}{HTML}{D0D7DE}

\DefineVerbatimEnvironment{PromptVerbatim}{Verbatim}{
  breaklines=true,
  breakanywhere=true,
  fontsize=\footnotesize,
  frame=single,
  framesep=2mm,
  rulecolor=\color{promptframe},
  bgcolor=promptbg,
  bgcolorpadding=1mm
}
\begin{document}

\def\spacingset#1{\renewcommand{\baselinestretch}%
{#1}\small\normalsize} \spacingset{1}

\if1\anon
{
  \title{\bf Nonuniformity Principle in Human-AI Coworking}
  \author{An Luo and Jie Ding\\
    School of Statistics, University of Minnesota\\
luo00318@umn.edu and dingj@umn.edu}
\date{}
  \maketitle
} \fi

\if0\anon
{
  \bigskip
  \bigskip
  \bigskip
  \begin{center}
    {\LARGE\bf 
    Nonuniformity Principle in Human-AI Coworking
    }
\end{center}
  \medskip
}
\bigskip
\fi
\begin{abstract}
As generative AI is increasingly applied to automate multi-step and high-stake workflows, human judgment and involvement remain essential for ensuring the quality of AI-generated outputs. In practice, while it is
desirable for human experts to provide oversight on AI regularly, often by reviewing intermediate outputs, giving feedback, making corrections, and steering subsequent steps, such oversight is constrained by the time and resources that humans can
afford. This creates a tension between the need for human oversight and AI's efficiency
in delivering more output with less intervention. An important but underexplored
question, then, is how to optimally engage humans in human-AI coworking. This work was originally motivated by our empirical observation that in long AI workflows, human oversight often improves user satisfaction while reducing unnecessary rework and token consumption. From there, we formulate the problem of where to place oversight stages in human-AI coworking. Under reasonable assumptions, we then develop the nonuniformity principle, which states that the optimal schedule places oversight stages with non-decreasing gaps along the workflow.
 We empirically validate this principle in two common AI agent workflows: writing literature reviews and constructing websites. 
\end{abstract}

\noindent%
\vfill

\newpage
\spacingset{1.8} % DON'T change the spacing!

\section{Introduction}
Generative AI is increasingly moving from single-shot generation based on large language models (LLMs)~\citep{wei2022chain, ouyang2022training, openai2023gpt4, anil2023gemini} toward long-horizon workflows, such as resolving real-world software engineering issues \citep{jimenez2024swebench}, navigating the web to accomplish user-specified goals \citep{he2024webvoyager},
and producing extended written reports \citep{wang2023survey}. In such long-horizon workflows, AI needs to work over multiple steps~\citep{yao2023react}, use external tools~\citep{qin2024toolllm,patil2024gorilla}, and coordinate different operations~\citep{hong2024metagpt,wu2024autogen}. It remains critical, however, to keep human oversight in the loop. For example, when AI is used to automate drug discovery~\citep{koscher2023autonomous,abramson2024alphafold3,demeo2025activelearning}, human experts still need to engage at multiple stages of the workflow, such as refining the biological objective, assessing whether proposed candidates are scientifically meaningful, and deciding which ones should be further experimentally validated. When AI automates laboratory operation~\citep{boiko2023autonomous,szymanski2023alab,dai2024autonomous}, humans still need to provide oversight at multiple stages, such as specifying experimental constraints, monitoring safety, and judging whether the measurements support the intended claim. 
% When AI works as a co-scientist~\citep{gottweis2026coscientist}, humans still need to judge the quality of generated hypotheses, check the evidence, and decide which directions are worth pursuing. 

In practice, while it is
desirable for human experts to provide oversight on AI regularly to ensure the quality of its output, such oversight is constrained by the time and resources that humans can
afford. This creates a tension between the need for human oversight and AI's efficiency
in delivering more output with less intervention. An important but underexplored
question, then, is how to optimally engage humans in human-AI coworking.
Existing research provides limited guidance on this question. Much work has studied how humans should provide feedback to AI systems \citep{ amershi2019guidelines,ouyang2022training}, while a growing literature examines the benefits of human involvement in complex domain-specific tasks, including medical decision-making \citep{reverberi2022experimental, vaccaro2024combinations, wang2026llm_clinical}, scientific writing \citep{gero2022sparks, liang2024llmfeedback, thakkar2026randomized}, and data science \citep{Meng2023, luo2025agentds, Luo2025AssistedDSBH,luo2026agentds}. Much less is known, however, about how human oversight with a limited number of human oversight stages should be scheduled within a long-horizon workflow of AI.

\begin{figure}[htbp]
\centering
\includegraphics[width=0.9\textwidth]{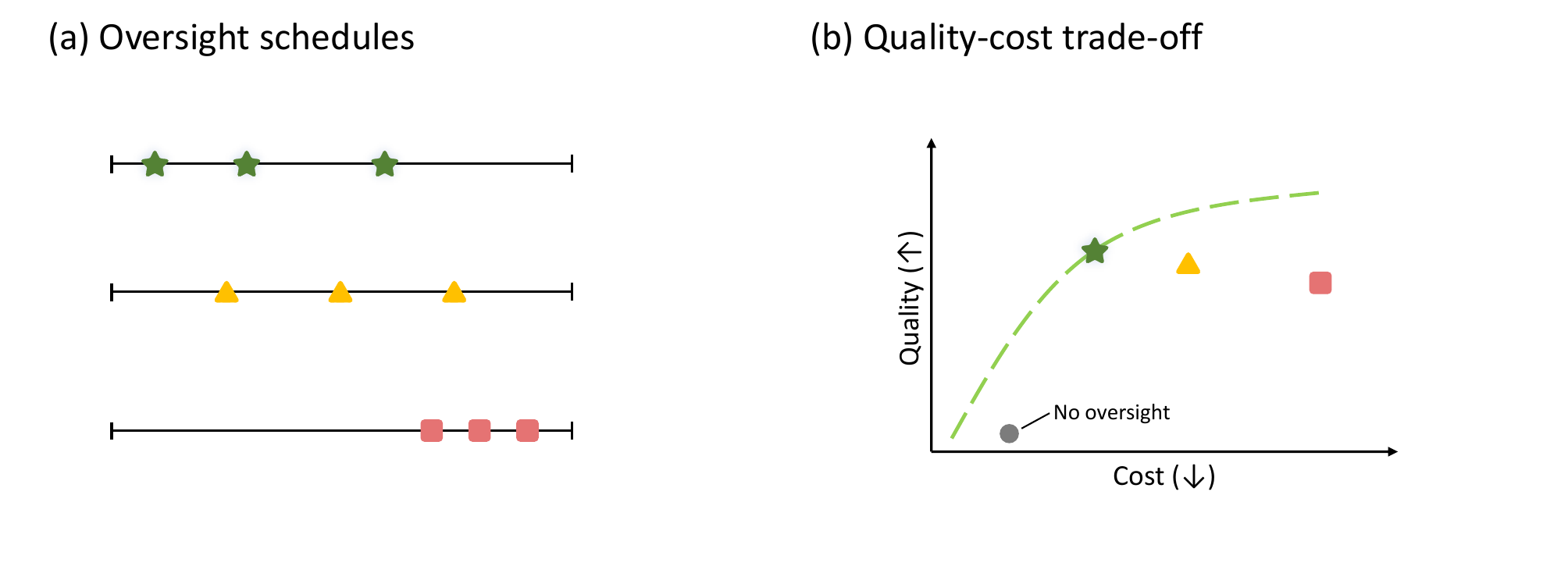}
\caption[]{Illustration of the nonuniformity principle. (a) Different oversight schedules with the same number of oversight stages.
The \textcolor{green!80!black}{green star} schedule places oversight relatively densely near the beginning and then uses increasing gaps between later oversight stages, which follows the nonuniformity principle.
The \textcolor{yellow!50!orange}{yellow triangle} schedule uses uniform gaps, and the \textcolor{red!90!magenta}{red square} schedule uses decreasing gaps.
(b) Quality-cost trade-off for these schedules. Here, cost stands for the human oversight cost.
The schedule under the nonuniformity principle is optimal among all four schedules.
The uniform and decreasing-gap schedules require larger cost without improving quality.
The \textcolor{gray}{gray circle} represents no oversight, which has the lowest cost but also the lowest quality.}
\label{fig:overview}
\end{figure}

Our investigation is motivated by the empirical observation that in long AI workflows, human oversight often improves user satisfaction while reducing unnecessary rework and token consumption. From there, we formulate the problem of where to place the oversight stages in human-AI coworking. Under reasonable assumptions, we then develop the \emph{nonuniformity principle}, which states that the optimal schedule places oversight stages with non-decreasing gaps along the workflow. Figure~\ref{fig:overview} gives an illustration of the nonuniformity principle.

We first formulate the problem of human-AI coworking, where an AI agent builds a deliverable step by step while the human's intention of what the deliverable should ultimately satisfy remains hidden from the agent.
The agent only starts with an initial context and must produce the final deliverable in $T$ stages. At selected stages, the human provides oversight based on the underlying intention and what the agent already produced, and the agent can revise the deliverable produced so far based on the human input. An oversight cost is incurred by the human in these stages. For a fixed number $K$ of oversight stages, the goal is to optimize the schedule of the oversight stages $S=\{s_1,\ldots,s_K\}$ to balance two forces: the quality of alignment between the final deliverable produced by the agent and the human intention, and the human oversight cost.

Building upon our formulation of human-AI coworking, the key idea behind our theory is to measure what happens between two consecutive oversight stages. We assume that after the human provides oversight, the agent is better aligned with the human intention. As the agent then works on its own for more stages, its uncertainty about the human intention can grow, so the expected alignment error is assumed to increase with the number of stages since the last oversight. 
% Under these assumptions, the original scheduling problem can be reduced to a much simpler form, which is only associated with the gaps between the $K$ oversight stages. Each gap contributes an accumulated alignment loss, and each oversight stage adds a human oversight cost. 
Under reasonable assumptions, we will show that the original scheduling problem can be reduced to a much simpler form, which pertains to scheduling between the $K$ oversight stages. 
And we further develop the nonuniformity principle that the optimal oversight schedules have non-decreasing gaps. Here, a gap means the number of production stages between neighboring human oversight. The resulting schedule uses oversight more frequently early on. Intuition is that, at early stages, human oversight can quickly narrow the AI’s long-term search space to align with human's unobserved intent. Later in the process, oversight becomes more costly but is still necessary to continue to refine the work to deliver a high-quality final result. We demonstrate the practical  value of the nonuniformity principle through experiments on two common long-horizon tasks: writing literature reviews and constructing HTML pages. 

The remainder of the paper is organized as follows.
Section~\ref{sec:formulation} formalizes the problem of human-AI coworking.
Section~\ref{sec:oversight-principle} develops the nonuniformity principle and provides a practical guide to find the optimal oversight schedule.
Section~\ref{sec:experiments} presents experimental results and examines their agreement with the theory. We conclude this paper in Section~\ref{sec:conclusion}. Supplementary material includes proofs and details of discussions and experiments.

%=======================================================================
%=======================================================================
\section{Problem Formulation of Human-AI Coworking}\label{sec:formulation}
\begin{figure}[htbp]
\centering
\includegraphics[width=1\textwidth]{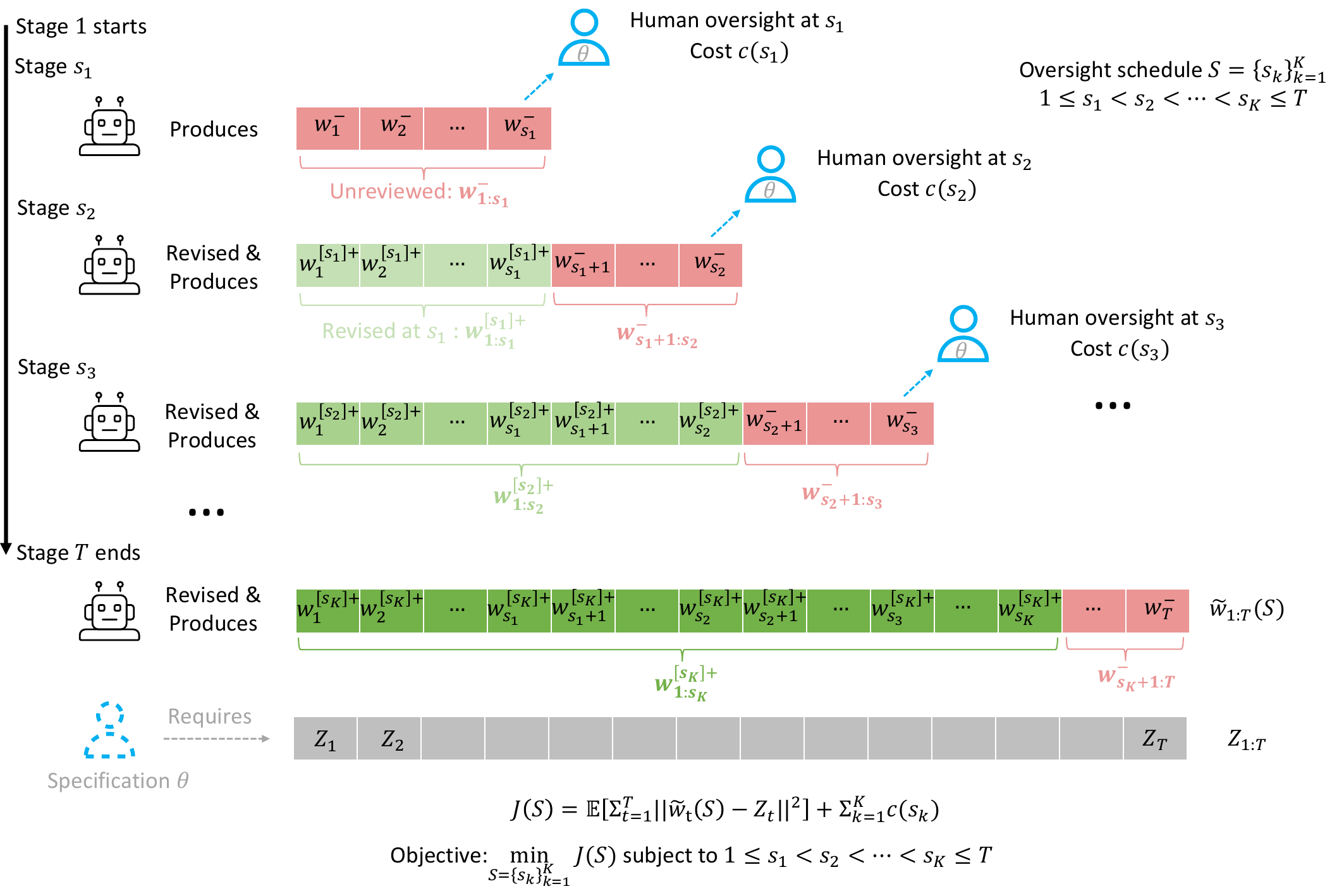}
\caption[]{
Overview of main concepts in human-AI coworking. 
The AI agent constructs a deliverable sequentially from stage \(1\) to stage \(T\). 
At each oversight stage \(s_k\) in the schedule
\(S=\{s_k\}_{k=1}^K\), with \(1\le s_1<s_2<\cdots<s_K\le T\), 
the human provides oversight based on the specification \(\theta\), incurring oversight cost \(c(s_k)\). 
Red blocks denote newly produced but not yet reviewed components \(w_t^-\), while green blocks denote components revised after human oversight at $s_k$, such as \(w_{1:s_k}^{[s_k]+}\). 
Thus, between two consecutive oversight stages, the agent continues producing unreviewed components, and at the next oversight the human reviews the current deliverable according to \(\theta\), after which the agent revises the reviewed components and continues production. 
After the final oversight stage \(s_K\), the remaining components \(w^-_{s_K+1:T}\) are produced without further review. 
The resulting final deliverable under schedule \(S\) is denoted by \(\widetilde w_{1:T}(S)\). 
The same specification \(\theta\) also determines the stage-level requirements \(Z_{1:T}\), against which the final deliverable is evaluated. 
The scheduling objective is to choose \(S\) to balance final alignment loss and human oversight cost:
$J(S)
=
\mathbb{E}\!\left[
\sum_{t=1}^T \|\widetilde w_t(S)-Z_t\|^2
\right]
+
\sum_{k=1}^K c(s_k),$
subject to \(1\le s_1<s_2<\cdots<s_K\le T\).
}
\label{fig:formulation}
\end{figure}

%=======================================================================
We begin with a description of the human-AI coworking problem.
A human has an intended deliverable in mind, but this intent is only partially available to the AI agent through the initial context.
Starting from this initial context, the agent constructs the deliverable over $T$ sequential stages, producing one component at a time.
At selected stages, the human reviews the partial deliverable produced so far and provides oversight.
Such oversight can help revise previously produced content, clarify the human's intent, and guide the agent's future production. At such oversight stages, the agent revises the working deliverable and then proceeds.
The final deliverable is evaluated by how well each stage-level output aligns with the corresponding latent requirement implied by the human's intention.
Each oversight also incurs a human oversight cost, as the human must spend time and effort inspecting the current draft before giving feedback.
The goal is to schedule a fixed number of oversight stages so that the final deliverable has high alignment quality and the human oversight cost remains low.
 An overview of main concepts in the formulation of human-AI coworking is given in Figure~\ref{fig:formulation}.

In this paper, an AI agent, or simply an agent, refers to a system that integrates data, tools, memory, operations, and human feedback to continuously generate actions~\citep{tian2025outlook}. We consider an AI agent that carries out the work through up to $T$ production stages. If a task has fewer than $T$ stages, one may add dummy stages that produce no new substantive content.
Without loss of generality, we suppose the agent's work consists of $T$ production stages.
In some tasks, the stages are natural production units.
For example, writing a paper may be organized into $T=6$ parts, such as abstract, introduction, related work, method, experiments, and conclusion.
In some other tasks, the stages may be milestones in a pipeline.
For example, a data analysis task may proceed through stages such as data cleaning, exploratory analysis, model fitting, validation, and report writing.

We suppose the human has an intended deliverable in mind when coworking with AI.
We denote this intended deliverable by a specification $\theta\in\Theta$, where $\Theta$ is the space of possible specifications. 
The specification $\theta$ determines what the human would regard as correct, complete, and well aligned with the task. The AI agent does not observe $\theta$, as the full specification may be highly dependent on domain knowledge and too costly to communicate before production begins.
For example, in scientific writing, $\Theta$ may include the intended argument, the relevant literature, the desired level of technical detail, and the author's judgment about what should be emphasized. Such information can be costly to write down in full and may involve domain knowledge that is difficult for the agent to infer from the initial description alone.
What is available to the AI agent is an initial context $D\in\mathcal D$, where $\mathcal D$ is the space of possible initial contexts. $\mathcal D$ is a general set that can include the task description, examples, available tools, data sources, reference materials, and other resources that the agent can use when producing the deliverable.

 At each stage, there is a corresponding target requirement implied by $\theta$. It is what the current component should accomplish in order for the final deliverable to match the human's intent. For example, in paper writing, the introduction should motivate the problem, the related work should position the paper against prior studies, and the method section should explain the proposed approach. 
Let $Z_t\in\mathcal Z$ denote the requirement at stage $t$, where $\mathcal Z$ is the requirement space. For technical simplicity, we set $\mathcal Z=\mathcal W$, treating the requirement at each stage as the ideal deliverable for that stage.
Let $Z_{1:T}=(Z_1,\ldots,Z_T)$ denote the full sequence of stage-level requirements. We also introduce $Z_0$, a latent initial state representing the requirement before production begins.
Let $Q_{\theta,0}$ denote the distribution of $Z_0$. Conditional on $\theta$, we model $Z_{1:T}$ as a general conditional process,
\begin{equation}
Z_t\mid (Z_{0:t-1}=z_{0:t-1},\theta)
\sim
Q_{\theta}(\cdot\mid z_{0:t-1}),
\qquad t=1,\ldots,T,
\label{eq:local-requirement-process}
\end{equation}
where $Q_{\theta}(\cdot\mid z_{0:t-1})$ is the conditional distribution of $Z_t$ given the past requirements $z_{0:t-1}$ and specification $\theta$.

At each stage $t=1,\ldots,T$, the agent produces a draft $w^-_t\in \mathcal W$, not yet reviewed. Depending on the task, $w^-_t$ may be a paragraph, a code section, a table, or another task-specific component. The space of deliverables across all $T$ stages is $\mathcal W^T$.

The human provides oversight at $K$ selected stages. Oversight may take different forms: clarifying intent, correcting content, giving feedback on the partial deliverable, or providing task-specific evidence such as test or execution results. An oversight schedule is a set $S:=\{s_k\}_{k=1}^K$ with $1\le s_1<\cdots<s_K\le T$ and $K<T$, since the human cannot review every stage. Each oversight stage $s\in S$ incurs a cost $c(s)\geq 0$, reflecting the effort to inspect the partial deliverable at stage $s$ and give feedback.

At an oversight stage $s\in S$, the agent's deliverable has two parts: the revised deliverable from the last oversight, $w^{[\tau(s)]+}_{1:\tau(s)}\in \mathcal W^{\tau(s)}$ (with $w^{[0]+}_{1:0}=\varnothing$ at the first oversight), and the new drafts $w^-_{\tau(s)+1:s}\in \mathcal W^{s-\tau(s)}$ produced since then. Here $\tau(t) := \max\bigl(\{0\}\cup\{s_k\in S:s_k<t\}\bigr)$ is the most recent oversight stage before $t$; $\tau(t) = 0$ means no prior oversight. Based on $\theta$,
the human reviews the current deliverable and returns feedback
$Y_s:
=
O_s\bigl(\theta,\,(w^{[\tau(s)]+}_{1:\tau(s)},w^-_{\tau(s)+1:s})\bigr)
\in\mathcal Y,$
where $O_s:\Theta\times\mathcal W^s\to\mathcal Y$ is the feedback operator. The feedback space $\mathcal Y$ is general: depending on the task, an element of $\mathcal Y$ may be natural language, execution results from external tools, or other task-specific information.
The feedback may suggest revisions to the current content and guide the agent's remaining stages. Based on $Y_s$, the agent revises the current deliverable and
produces
$w^{[s]+}_{1:s}
=
R_s\bigl(Y_s,\,(w^{[\tau(s)]+}_{1:\tau(s)},w^-_{\tau(s)+1:s})\bigr)
\in\mathcal W^s,$
where $R_s:\mathcal Y\times\mathcal W^s\to\mathcal W^s$ is the revision operator. The agent also maintains a memory of the feedback
$M_s:=\bigl(Y_r:r\in S,r\le s\bigr)$
at stage $s\in S$, with $M_0=\varnothing$.

The agent produces $w^-_t \in \mathcal W$ based on
$\mathcal H_{t-1}
:=
\bigl(D,\;w^{[\tau(t)]+}_{1:\tau(t)},\;w^-_{\tau(t)+1:t-1},\;M_{\tau(t)}\bigr)$,
which comprises the initial context $D$, the revised deliverable from the last oversight $\tau(t)$, the drafts produced since then, and the accumulated feedback $M_{\tau(t)}$.
At an oversight stage $s\in S$, no new drafts exist yet at $s+1$, so $\mathcal H_s = \bigl(D,\;w_{1:s}^{[s]+},\;M_s\bigr)$.
For the theoretical analysis, we model the agent's actions as following the Bayes decision rule:
\begin{equation}
\label{eq:bayes-agent-action}
w_t^-
\in
\arg\min_{w\in\mathcal{W}}
\mathbb{E}_{Z_t \mid
\mathcal{H}_{t-1}}\!\left[
\ell(w,Z_t)
\right].
\end{equation}
To develop technical results, we consider the case $\mathcal{W}=\mathcal{Z}=\mathbb{R}$ with $\ell(w,Z_t) =\|w-Z_t\|^2 $, where $\|\cdot\|$ is the Euclidean norm. Under this loss function, rule~\eqref{eq:bayes-agent-action} gives $w_t^- = \mathbb{E}[Z_t\mid\mathcal{H}_{t-1}]$. In the experimental studies, we will consider general loss functions.

% Section~\ref{sec:oversight-principle} derives the scheduling objective from these model objects and characterizes its optimal solution.

Let $\widetilde{w}_{1:T}(S)=(\widetilde{w}_1(S),\ldots,\widetilde{w}_T(S))$ denote the final deliverable under oversight schedule $S$. This is the fully revised deliverable following the procedure above.
For each stage $t$, $\widetilde{w}_t(S)$ is given by
\begin{equation}
\label{eq:final-output}
\widetilde{w}_t(S)
:=
\begin{cases}
w_t^{[s_K]+} & \text{if } t\leq s_K,\\
w_t^- & \text{if } s_K < t\le T.
\end{cases}
\end{equation}

Define the expected alignment loss under $S$ as
\begin{equation}
\label{eq:final-loss}
L(S)
: =
\mathbb{E}\left[
\sum_{t=1}^T \|\widetilde{w}_t(S)-Z_t\|^2\right].
\end{equation}
Define
$J(S)
:=
L(S)+\sum_{k=1}^K c(s_k)$ as the total loss, combining alignment loss and oversight cost. The objective is
\begin{equation}
\label{eq:obj}
\min_{S=\{s_k\}^K_{k=1}}
J(S)
\quad
\text{s.t.}
\quad
1\leq s_1<\cdots<s_K\leq T.
\end{equation}
It aims to find the schedule $S$ that best balances alignment quality and oversight cost.

%=======================================================================
\section{The Nonuniformity Principle}
\label{sec:oversight-principle}
%=======================================================================
In this section, we develop the nonuniformity principle. In Section~\ref{subsec:block-loss}, we introduce some assumptions. In Section~\ref{subsec:nonuniform}, we present nonuniformity principle as the main results. In Section~\ref{subsec:linear-cost}, we provide a practical guide for finding the optimal schedule.
% Between two successive review stages, the agent works autonomously for $d_j$
% stages --- we call this a \emph{block} of length $d_j$, where $j$ indexes the block
% ($d$ is used generically when block index is immaterial).

%-----------------------------------------------------------------------
\subsection{Assumptions and preparations}
\label{subsec:block-loss}

Suppose after an oversight stage $s\in S$, the agent produces $d\geq 1$ additional drafts $w^-_{s+1},\ldots,w^-_{s+d}$
without yet being reviewed, i.e., $\tau(s+d) =s$. To measure how prediction error accumulates after an
oversight, let $\rho_s(r)$ denote the expected
conditional variance of $Z_{s+r}$ given the information available at stage $s$, i.e.,
\begin{equation}
\label{eq:lag-risk-stage-specific}
\rho_s(r)
:=
\mathbb{E}\!\left[
\operatorname{Var}\!\left(
Z_{s+r}
\mid
\mathcal{H}_s
\right)
\right],
\qquad r=1,\ldots,d.
\end{equation}

\begin{assumption}
\label{ass:ci}
For each $s\in S\cup\{0\}$ and each integer $r>1$, $Z_{s+r}$ is conditionally independent of
$(w_{s+1}^-,\ldots,w_{s+r-1}^-)$ given $\mathcal{H}_s$.
\end{assumption}

Assumption~\ref{ass:ci} decouples the latent requirement process from the agent's
intermediate deliverables. That is, the future requirement $Z_{s+r}$ depends only on the information available at stage $s$, not on the drafts produced in between.

\begin{lemma}
\label{lem:bayes-mse}
Under Assumption~\ref{ass:ci}, for each $s\in S\cup\{0\}$ and each positive integer $r$ satisfying $\tau(s+r) = s$,
\begin{equation}\label{eq:eq_rho}
  \mathbb{E}\!\left[\|w_{s+r}^- - Z_{s+r}\|^2\right] = \rho_s(r).  
\end{equation}

\end{lemma}

Lemma~\ref{lem:bayes-mse} says that the expected squared error of the agent's draft at stage $s+r$ equals $\rho_s(r)$, the conditional variance of $Z_{s+r}$ as seen from stage $s$.

\begin{assumption}
\label{ass:lag-risk-homogeneity}
There exists a function $\rho(r)\ge 0$ such that
$\rho_s(r)=\rho(r)$ for any $s\in S\cup\{0\}$ and any integer $r\ge 1$ satisfying $\tau(s+r)=s$.
\end{assumption}

Assumption~\ref{ass:lag-risk-homogeneity} states that, after any oversight, the expected prediction error at lag $r$ depends only on $r$, not on which stage the oversight occurs.

\begin{remark}[When does Assumption~\ref{ass:lag-risk-homogeneity} hold?]
\label{rem:lag-risk-homogeneity}
Suppose the oversight at stage $s$ is clear enough to fully resolve the current requirement, meaning $\operatorname{Var}(Z_s\mid \mathcal H_s)=0$ for each $s\in S\cup\{0\}$. Then all remaining uncertainty about $Z_{s+r}$ comes from future evolution alone. Under this condition, we give two examples in which Assumption~\ref{ass:lag-risk-homogeneity} holds. 
\textbf{1) Random walk.}
$Z_{t+1}=Z_t+\varepsilon_{t+1}$ for $t=0,\ldots,T-1$, where the innovations $\varepsilon_1,\ldots,\varepsilon_T$ are independent, have mean zero, and have common variance $\sigma^2>0$. Then, for any $s\in S\cup\{0\}$ and any $r\ge 1$ satisfying $\tau(s+r)=s$, we have $Z_{s+r}=Z_s+\sum_{q=1}^r \varepsilon_{s+q}$, and $\operatorname{Var}(Z_{s+r}\mid \mathcal H_s)=\sum_{q=1}^r \operatorname{Var}(\varepsilon_{s+q})=\sigma^2 r$. Therefore $\rho_s(r)=\rho(r)=\sigma^2 r$.
\textbf{2) Stable AR(1).}
$Z_{t+1}=\mu+\alpha(Z_t-\mu)+\varepsilon_{t+1}$ for $t=0,\ldots,T-1$, where $\mu$ is a constant, $0<|\alpha|<1$, and the innovations $\varepsilon_1,\ldots,\varepsilon_T$ are independent, have mean zero, and have common variance $\sigma_\varepsilon^2>0$. Then, for any $s\in S\cup\{0\}$ and any $r\ge 1$ satisfying $\tau(s+r)=s$, we have $Z_{s+r}=\mu+\alpha^r(Z_s-\mu)+\sum_{q=1}^r \alpha^{r-q}\varepsilon_{s+q}$ and $\operatorname{Var}(Z_{s+r}\mid \mathcal H_s)=\sigma_\varepsilon^2\sum_{q=1}^r \alpha^{2(r-q)}=\sigma_\varepsilon^2(1-\alpha^{2r})/(1-\alpha^2)$. Therefore $\rho_s(r)=\rho(r)=\sigma_\varepsilon^2(1-\alpha^{2r})/(1-\alpha^2)$. In both examples, Assumption~\ref{ass:lag-risk-homogeneity} holds.
\end{remark}

Together, Lemma~\ref{lem:bayes-mse} and Assumption~\ref{ass:lag-risk-homogeneity} establish that 
$\mathbb{E}[\|w_{s+r}^- - Z_{s+r}\|^2] = \rho_s(r) = \rho(r)$.

With $s_0:=0$, we impose the following assumption on the final deliverable.

\begin{assumption}
\label{ass:contraction}
There exists a constant $\kappa\in(0,1)$ such that, for every $k=0,\ldots,K-1$ and every integer $t$ satisfying $s_k<t\le s_{k+1}$, where $s_1,\ldots,s_K$ are the oversight stages in $S$,
$\mathbb E\!\left[
\|\widetilde w_t(S)-Z_t\|^2
\right]
=
\kappa\,\rho(t-s_k).$
\end{assumption}

Assumption~\ref{ass:contraction} states that the expected squared error at any stage between two oversights is a fixed fraction $\kappa$ of $\rho$. This captures the benefit of oversight: reviewed stages have lower error (by factor $\kappa < 1$) than they would without it.

We can now decompose $L(S)$ in terms of the gaps between oversight stages. By Lemma~\ref{lem:bayes-mse} and Assumption~\ref{ass:lag-risk-homogeneity},
we have
$\mathbb{E}[\|w_{s+r}^- - Z_{s+r}\|^2] = \rho_s(r) = \rho(r)$ for any $s\in S\cup\{0\}$ and for each $r=1,\ldots,d$.
By Assumption~\ref{ass:contraction}, we have
$\mathbb{E}[\|\widetilde{w}_{s+r}-Z_{s+r}\|^2] = \kappa\,\rho(r)$ for any $s\in \{0\} \cup S\setminus \{s_K\}$ and for each $r=1,\ldots,d$. So for any $s\in \{0\} \cup S\setminus \{s_K\}$ we have $\sum_{r=1}^d \mathbb{E}\!\left[\|\widetilde{w}_{s+r}-Z_{s+r}\|^2\right]
=
\kappa\sum_{r=1}^d\rho(r)
: =
\Phi(d)$ for $d\geq1$, and we set $\Phi(0) = 0$. For $s=s_K$ we have $\sum_{r=1}^d \mathbb{E}\!\left[\|\widetilde{w}_{s+r}-Z_{s+r}\|^2\right]
=
\sum_{r=1}^d\rho(r)
: =
\Psi(d)$ for $d\geq 1$, and we set $\Psi(0) = 0$.

\begin{assumption}
\label{ass:staleness}
$\rho(r)$ is strictly increasing in $r$.
\end{assumption}

\begin{remark}[When does Assumption~\ref{ass:staleness} hold?]
\label{rem:staleness}
We consider the same two models in Remark~\ref{rem:lag-risk-homogeneity}:
\textbf{1) Random walk.} $\rho(r)=\sigma^2 r$ is strictly increasing in $r$.
\textbf{2) Stable AR(1).} $\rho(r)=\sigma_\varepsilon^2(1-\alpha^{2r})/(1-\alpha^2)$ is strictly increasing in $r$.
In both cases, Assumption~\ref{ass:staleness} holds.
\end{remark}
\begin{assumption}
\label{ass:increasing-cost}
The oversight cost $c(s)$ is strictly increasing in $s$.
\end{assumption}

\begin{lemma}
\label{lem:convexity}
Under Assumption~\ref{ass:staleness}, $\Phi(d)$ and $\Psi(d)$ are strictly
increasing and strictly discrete convex in $d$.
\end{lemma}

Let $s_{K+1} := T$, and $d_k := s_{k+1}-s_k$ for $k=0,\ldots,K$. Since $L(S) = \mathbb{E}\left[
\sum_{t=1}^T \|\widetilde{w}_t(S)-Z_t\|^2\right] = \sum_{t=1}^T \mathbb{E}\left[
\|\widetilde{w}_t(S)-Z_t\|^2\right] = \sum_{k=0}^{K}
\sum_{t=s_k+1}^{s_{k+1}}
\mathbb{E}\left[
\|\widetilde{w}_t(S)-Z_t\|^2
\right]=  \sum_{k=0}^{K}\sum_{r=1}^{d_k}
\mathbb{E}\left[
\|\widetilde{w}_{s_k + r}(S)-Z_{s_k + r}\|^2 
\right] = \sum_{j=0}^{K-1}\Phi(d_j) + \Psi(d_K)$,
the objective of minimizing the loss~\eqref{eq:final-loss} (but without oversight costs) reduces to
\begin{equation}
\label{eq:cost-free-objective}
\min_{d_0,\ldots,d_K \in \mathbb Z}
\sum_{j=0}^{K-1}\Phi(d_j) +\Psi(d_K)
\quad
\text{s.t.}
\quad
\sum_{j=0}^K d_j=T, d_0,\ldots,d_{K-1}\geq 1,d_K\ge 0.
\end{equation}

The following Proposition~\ref{prop:uniform} shows that when the oversight cost is negligible, the optimal schedule spreads nearly uniformly: any two gaps between consecutive oversight stages differ by at most one.

\begin{proposition}
\label{prop:uniform}
Let $\bm d^{\star,0}:=(d_0^{\star,0},\ldots,d_K^{\star,0})$ be any minimizer of~\eqref{eq:cost-free-objective}. Under Assumption~\ref{ass:staleness},
$\max_{0\le j\le K-1}d_j^{\star,0}-\min_{0\le j\le K-1}d_j^{\star,0}\le 1.$
% Moreover, for any $j=0,\ldots,K-1$ with $d_j^{\star,0}\ge 2$, $\kappa\rho(d_j^{\star,0})\le \rho(d_K^{\star,0}+1),$
% and, if $d_K^{\star,0}\ge 1$, then for every $j=0,\ldots,K-1$, $\rho(d_K^{\star,0})\le \kappa\rho(d_j^{\star,0}+1).$
\end{proposition}

%-----------------------------------------------------------------------

\subsection{Non-decreasing gaps under increasing oversight cost}
\label{subsec:nonuniform}
We now consider the objective~\eqref{eq:obj} for quality-cost trade-off.  
Using the decomposition
$L(S)=\sum_{j=0}^{K-1}\Phi(d_j)+\Psi(d_K)$, the objective with oversight cost, objective~\eqref{eq:obj}, becomes
\begin{equation}
\label{eq:general-objective}
\min_{d_0,\ldots,d_K  \in \mathbb Z}
\sum_{j=0}^{K-1}\Phi(d_j)
+
\Psi(d_K)
+
\sum_{k=1}^K c(s_k)
\quad
\text{s.t.}
\quad
\sum_{j=0}^K d_j=T,d_0,\ldots,d_{K-1}\geq 1,d_K\ge 0,
\end{equation}
where $s_k=\sum_{l=0}^{k-1}d_l.$

\begin{remark}[Existence of the optimal schedule]
\label{rem:discrete-existence}
The feasible set in \eqref{eq:general-objective} is finite, so a global minimizer exists. 
\end{remark}

\begin{theorem}[The nonuniformity principle: non-decreasing gaps]
\label{thm:nonuniformity}
Under Assumptions~\ref{ass:staleness} and~\ref{ass:increasing-cost}, every minimizer
${\bm d}^\star:=(d_0^\star,\ldots,d_K^\star)$ of \eqref{eq:general-objective} satisfies
$d_0^\star\le d_1^\star\le\cdots\le d_{K-1}^\star.$ 
\end{theorem}

Theorem~\ref{thm:nonuniformity} formalizes the nonuniformity principle: with a fixed number of oversight stages, an optimal schedule places oversight relatively densely early in the process, and the gaps before later oversight stages are no smaller than the earlier ones. Without oversight cost, Proposition~\ref{prop:uniform} shows that the reviewed
gaps differ no more than one.  The oversight cost breaks this
balance. As later oversight stages require reviewing a longer deliverable, the optimal
schedule shifts oversight stages earlier and produces gaps that are non-decreasing over
time.  

Theorem~\ref{thm:nonuniformity} does not impose a result involving $d_K^\star$. This is because $d_K^\star$ is the terminal gap after the last oversight stage, rather than a gap ending at an oversight stage. In applications, it is appealing to set $s_K=T$, equivalently $d_K=0$, and apply the same scheduling idea to the earlier oversight stages, as this corresponds to a final review-and-revision step after the agent has produced the final deliverable.

% The ordering concerns $d_0,\ldots,d_{K-1}$, the gaps whose right endpoints are
% oversight stages.  

\begin{theorem}[The nonuniformity principle: earliest oversight stages under high oversight cost]
\label{thm:all-early-general-cost}
Under Assumption~\ref{ass:staleness}, if
$\min_{1\le s\le T-1}\{c(s+1)-c(s)\}>\rho(T-K)-\kappa\rho(2),$
then 
$d_0^\star=\cdots=d_{K-1}^\star=1$ and $
d_K^\star=T-K.$
\end{theorem}
Theorem~\ref{thm:all-early-general-cost} states that, if the oversight cost grows fast, the optimal schedule would be to place all $K$ oversight stages at the first $K$ stages. This means that when it is too costly for the human to provide oversight, the oversight stages should be set as early as possible.

\begin{corollary}
\label{cor:all-early-linear-cost}
Suppose $c(s)=\lambda s$ with $\lambda>0$ in the objective~\eqref{eq:general-objective}.  Under Assumption~\ref{ass:staleness}, if
$\lambda>\rho(T-K)-\kappa\rho(2),$
then 
$d_0^\star=\cdots=d_{K-1}^\star=1$ and $
d_K^\star=T-K.$
\end{corollary}

%-----------------------------------------------------------------------

\subsection{A practical guide for finding the optimal schedule}
\label{subsec:linear-cost}

To give an exact algorithm for finding the optimal schedule as a practical guide, here we take
$c(s)=\lambda s$, where $\lambda >0$ is a constant representing how costly reviewing is to the human.
This choice reasonably assumes that reviewing a deliverable with $s$ units requires effort
proportional to the amount of content.  Since
$s_k=\sum_{j=0}^{k-1}d_j,$
we have
\begin{equation}\label{eq:linearcost}
\sum_{k=1}^K c(s_k)
=
\lambda\sum_{k=1}^K s_k
=
\lambda\sum_{k=1}^K\sum_{j=0}^{k-1}d_j
=
\lambda\sum_{j=0}^{K-1}(K-j)d_j.  
\end{equation}
To give a practically simple algorithm, here we assume that $Z_t$ evolves as the random walk model described in Remark~\ref{rem:lag-risk-homogeneity}. This gives $\rho(r)=\sigma^2 r$, and hence
\begin{equation}\label{eq:phiandpsi}
 \Phi(d)=\frac{\kappa\sigma^2}{2}d(d+1)
\text{ and }
\Psi(d)=\frac{\sigma^2}{2}d(d+1). 
\end{equation}

From the objective~\eqref{eq:general-objective}, with the random walk model under the linear oversight cost $c(s)=\lambda s$, the exact scheduling objective is given by (combining~\eqref{eq:general-objective},~\eqref{eq:linearcost}, and~\eqref{eq:phiandpsi})
\begin{equation}
\label{eq:rw-linear-discrete-objective}
J_\lambda(d_0,\ldots,d_K)
=
\sum_{j=0}^{K-1}
\left\{
\frac{\kappa\sigma^2}{2}d_j(d_j+1)
+
\lambda(K-j)d_j
\right\}
+
\frac{\sigma^2}{2}d_K(d_K+1),
\end{equation}
subject to $\sum_{j=0}^K d_j=T$, $d_0,\ldots,d_{K-1}\ge 1$, and $d_K\ge 0$.

Let
$\eta:={\lambda}/{\sigma^2}.$
Dividing \eqref{eq:rw-linear-discrete-objective} by $\sigma^2$ gives the normalized
objective
\begin{equation}
\label{eq:rw-linear-normalized-objective}
\bar J_{\kappa,\eta}(d_0,\ldots,d_K)
=
\sum_{j=0}^{K-1}
\left\{
\frac{\kappa}{2}d_j(d_j+1)
+
\eta(K-j)d_j
\right\}
+
\frac12 d_K(d_K+1),
\end{equation}
subject to $\sum_{j=0}^K d_j=T$, $d_0,\ldots,d_{K-1}\ge 1$, $d_K\ge 0$, and $d_0,\ldots,d_K\in \mathbb Z$.

The objective~\eqref{eq:rw-linear-normalized-objective} depends only on two effective
parameters: the revision factor $\kappa$ and the ratio
$\eta=\lambda/\sigma^2$. The following algorithm gives an exact schedule that is optimal.

\begin{algorithm}[htbp]
\caption{\label{alg:rw-schedule}
A scheduling guide based on the random walk model with linear cost}
\begin{algorithmic}[1]
\Require Number of production stages $T$, number of oversight stages $K$,
revision factor $\kappa\in(0,1)$,
ratio $\eta=\lambda/\sigma^2$.
\Ensure Gap vector $\widehat{\bm d}=(\widehat d_0,\ldots,\widehat d_K)$
and oversight schedule $\widehat S=\{\widehat s_1,\ldots,\widehat s_K\}$.

\State Initialize $\bm d = (d_0,\ldots,d_{K-1},d_K)\gets(1,\ldots,1,0)\in\mathbb Z^{K+1}$.
\State Set $B\gets T-K$.

\For{$b=1,\ldots,B$}
    \State Compute
    \[
    \Delta_j(\bm d)
    \gets
    \begin{cases}
    \kappa(d_j+1)+\eta(K-j), & j=0,\ldots,K-1,\\
    d_K+1, & j=K.
    \end{cases}
    \]
    \State Select any
    $j_b\in\arg\min_{j=0,\ldots,K}\Delta_j(\bm d).$
    \State Update $d_{j_b}\gets d_{j_b}+1$.
\EndFor

\State Set $\widehat{\bm d}\gets\bm d$.
\State Compute $\widehat s_k\gets\sum_{j=0}^{k-1}\widehat d_j$, for $k=1,\ldots,K$.
\State Set $\widehat S\gets\{\widehat s_1,\ldots,\widehat s_K\}$.
\State \Return $\widehat{\bm d}$ and $\widehat S$.
\end{algorithmic}
\end{algorithm}

\begin{proposition}
\label{prop:rw-schedule-algorithm}
Let \(T,K\in\mathbb Z^+\) with \(K<T\), \(\kappa\in(0,1)\), and \(\eta>0\). Then the gap vector \(\widehat{\bm d}\) returned by Algorithm~\ref{alg:rw-schedule} is a global minimizer of~\eqref{eq:rw-linear-normalized-objective}.
\end{proposition}

Algorithm~\ref{alg:rw-schedule} gives a direct implementation to find an optimal schedule guaranteed by Proposition~\ref{prop:rw-schedule-algorithm}.
The user needs to specify $T$, $K$, $\kappa$, and $\eta=\lambda/\sigma^2$. In practice, $T$ could be the number of natural production units, such as paragraphs in a writing task, sections in a webpage construction task, or modules in a coding task. The number of oversight stages $K$ is determined by how many times the human is willing or able to provide oversight. The revision factor $\kappa$ represents how much alignment error remains after human oversight. A small value of $\kappa$ corresponds to highly effective oversight, and a $\kappa$ closer to one corresponds to weaker oversight.  Thus, $\kappa\in (0,1)$ can be set based on how effective the user feels about the oversight they would provide. The parameter $\eta=\lambda/\sigma^2\in \mathbb R^+$ compares the burden of reviewing a longer deliverable with the uncertainty in the agent's production.  A smaller $\eta$ is appropriate when review is relatively easy, or when the user is more concerned about accumulated uncertainty and therefore willing to review later drafts. A larger $\eta$ is appropriate when reviewing longer drafts is burdensome, or when the user prefers to provide earlier oversight before the deliverable becomes costly to inspect.

\section{Experiments}
\label{sec:experiments}
%=======================================================================

In this section, we present our experimental observations that motivate the nonuniformity principle, and also verify that the theoretical results in Section~\ref{sec:oversight-principle} align with the experimental results. We consider two long-horizon tasks where the human specification is often not revealed in full: writing the related work section for a research paper, and constructing an HTML page. These two tasks represent two prominent contemporary applications of AI agents: text generation~\citep{Lin2025Creativity,Huot2025AgentsRoom} and code generation~\citep{jimenez2024swebench,Yang2024SWEagent,Wang2025OpenHands}.
The rest of this section is organized as follows: Section~\ref{subsec:shared} presents the shared experimental setup for the two tasks. Section~\ref{subsec:study1} presents oversight scheduling for writing related work. Section~\ref{subsec:study2} presents the same problem in HTML page construction. Section~\ref{subsec:discussion} connects the empirical observations to the nonuniformity principle and the proposed algorithm in Section~\ref{sec:oversight-principle}.

\subsection{Setup shared by the two tasks}\label{subsec:shared}
For both tasks,  we set $T=10$ for production stages and $K=3$ for oversight stages.  To examine how different schedules behave in the quality-cost trade-off, we test six candidate schedules: five schedules whose gaps increase, remain flat, or decrease, and one schedule with no oversight serving as a baseline. Table~\ref{tab:schedules-all} presents details of the six schedules considered in our experiments. 

\begin{table}[htbp]
\centering
\small
\caption{Six oversight schedules considered in experiments ($T=10$, $K=3$ for all except
\textsc{Skip}). Oversight Schedule $S=\{s_k\}^3_{k=1}$ with $1\leq s_1<s_2<s_3\leq10$. Gaps $d_j = s_{j+1} - s_j$ with $s_0=0$ for $j=0,1,2$.}
\begin{tabular}{lccc}
\toprule
\textbf{Name} & \textbf{Oversight Schedule $S$} & \textbf{Gaps $(d_0,d_1,d_2)$} & \textbf{Gap trend} \\
\midrule
\textsc{Skip}        & ---     & ---             & --- \\
\textsc{Burst-Early} & $\{1,2,3\}$ & $(1,1,1)$       & flat \\
\textsc{Tilt-Early}  & $\{1,3,6\}$ & $(1,2,3)$       & increasing \\
\textsc{Spread}      & $\{1,4,9\}$ & $(1,3,5)$       & increasing \\
\textsc{Uniform}     & $\{2,5,8\}$ & $(2,3,3)$       & flat \\
\textsc{Burst-Late}  & $\{7,8,9\}$ & $(7,1,1)$       & decreasing \\
\bottomrule
\end{tabular}

\label{tab:schedules-all}
\end{table}

For each task we measure
a quality metric with range 1--10 (higher is better) and a cost metric (lower is better) and
identify the Pareto hull in the resulting quality-cost space. 

% The theory predicts that
% the hull should contain early-tilted, gap-increasing schedules while back-loaded and
% uniform schedules are dominated.

The objective~\eqref{eq:obj} defined in Section~\ref{sec:formulation} is a population expected loss over the latent
requirement process. In the experiments, each task provides one specification $\theta$ but the requirement process is still not directly available.  We therefore
estimate schedule-level performance by the empirical average over task instances.  We use the
quality score given by a judge as a proxy for the alignment quality, while the review
cost is measured directly from the length of the working deliverable. Details of how such a quality score is obtained are given in Sections~\ref{subsec:study1} and~\ref{subsec:study2}, respectively. For each schedule $S$, we report $\widehat Q(S)$, the average of the quality score over $N$ tasks, and $\widehat C(S)$, the average of the cost over $N$ tasks.
We use the loss ${\mathcal L}_{\lambda}(S)
=
10-\widehat Q(S) + \lambda \widehat C(S)$ as a proxy of the objective~\eqref{eq:obj} defined in Section~\ref{sec:formulation}. In this section, a schedule is on the Pareto hull if it minimizes
${\mathcal L}_{\lambda}(S)$ for some $\lambda\ge 0$ among the candidate
oversight schedules with $K=3$.

%-----------------------------------------------------------------------
\subsection{Study 1: writing related work}
\label{subsec:study1}

This study focuses on human-AI coworking for writing the related work section in a scientific paper.
We present the setup, metrics, and results below, and the full implementation details of this study are in Section~B of the supplementary material.

\textbf{Setup.}
We sample $N=40$ accepted papers from~\citet{iclr2026} stratified across 18 primary areas.
Related work sections in the papers sampled range from 203 to 658 words (median 281). For each paper as a task instance, we set up three roles, agent for writing related work, human for providing oversight, and judge for giving quality scores on the final deliverables, each realized with separate LLM calls with temperature $=0$: 
\begin{itemize}
\item \textbf{Agent} (writing related work, \texttt{deepseek-v4-flash}): given only the paper title
  and abstract, writes one paragraph of related work per step $t=1,\ldots,10$. Rewrite the deliverable at any oversight stage based on human oversight.
\item \textbf{Human} (providing oversight, \texttt{deepseek-v4-pro}): at any oversight stage $s\in S$, given the real related work section in the paper as specification $\theta$ and related work written by the agent up to stage $s$,
  returns feedback to the agent with suggestion for revision and guidance on writing the remaining parts.
\item \textbf{Judge} (giving quality score, \texttt{deepseek-v4-pro}): given the paper title, abstract, introduction, and the real related work as target, and the final deliverables from all six schedules, scores (1--10 integer scale) each deliverable on coverage and factual accuracy.
\end{itemize}

For each schedule listed in Table~\ref{tab:schedules-all}, we measure its quality through the judge, and its cost through the length of the deliverable, as explained in detail below.

\textbf{Quality.}
Two dimensions are scored by the judge:\textbf{coverage} (does the draft mention the key
prior works, methods, and findings?) and\textbf{factual accuracy} (are
citations, method names, and claims attested?). After all final deliverables are produced, for each paper $p$, the judge scores the deliverables produced under the six schedules. To reduce the effect of presentation order, we repeat this scoring three times, each time with a randomized order of the six deliverables presented to the judge. Then the medians across the three repetitions are set as the quality score for each dimension.
The paper-level quality per schedule is then given by
the average of the two dimensions' quality scores.

\textbf{Cost.}
Human oversight cost is measured by the total length of draft content the human must read when reviewing. To obtain it, each time the LLM with human role is called we measure the length of the draft itself. The paper-level cost per schedule is then given by the sum of such lengths at the schedule's oversight stages.

\begin{figure}[htbp]
\centering
\includegraphics[width=1\textwidth]{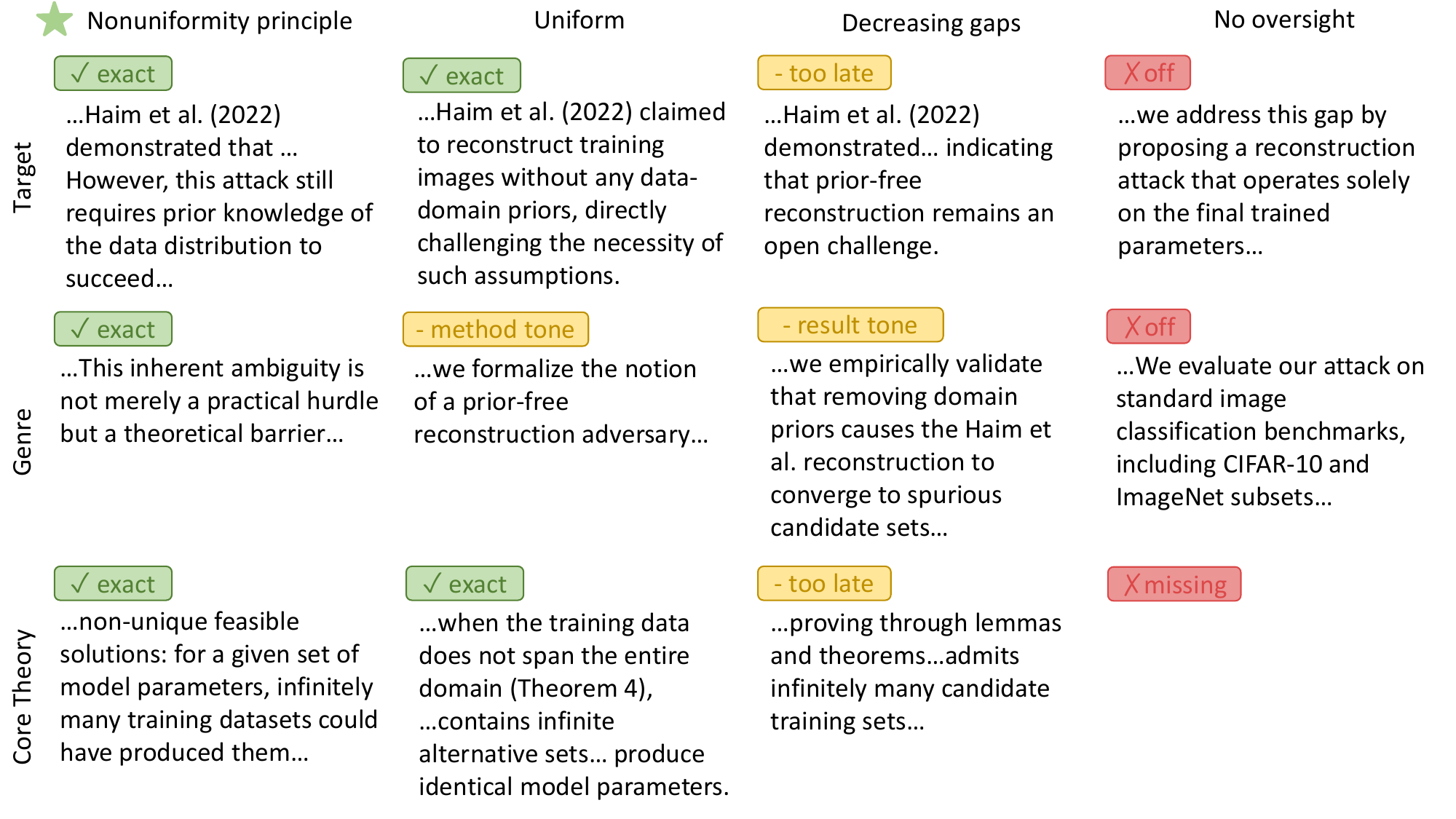}
\caption[]{Qualitative comparison of one Study~1 example, showing the final related work sections produced under four distinct oversight schedules.
Each column corresponds to one schedule, and each row shows one aspect of the draft.
The schedule satisfying the nonuniformity principle (\textsc{Spread}, with increasing gaps) performs best: it correctly identifies the Haim et al. attack as the target, keeps the text in a related work tone, and captures the core non-uniqueness argument as theory.
The schedule with uniform gaps (\textsc{Uniform}) also identifies the target and includes some core theory, but it drifts toward the method tone.
The schedule with decreasing gaps (\textsc{Burst-Late}) partially recovers the target and the core theory but writes too late in the draft, and it retains a result tone.
The schedule with no oversight (\textsc{Skip}) performs worst: it changes target by describing a new reconstruction attack and omits the core theory.}
\label{fig:study1comp}
\end{figure}

\begin{figure}[htbp]
\centering
\includegraphics[width=0.85\textwidth]{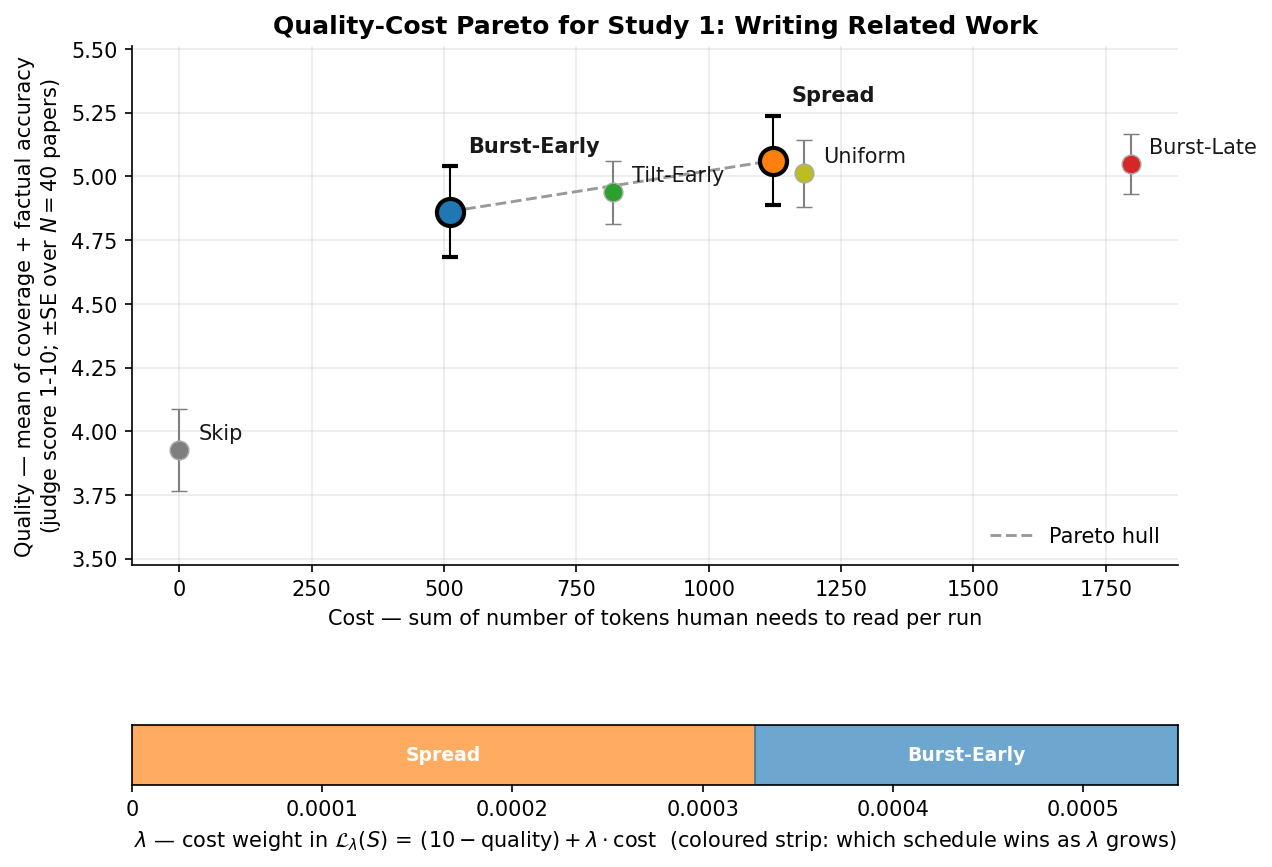}
\caption[]{Quality-cost Pareto frontier of study~1: writing related work. Summarized on $N=40$
ICLR~2026 papers. Each point is one oversight schedule. Error bars show
$\pm 1$~SE across papers. \textbf{Quality}
(vertical axis): mean composite judge score (mean of median coverage and factual
accuracy, 1--10 scale). \textbf{Cost} (horizontal axis): mean variable human
read tokens per paper, intercept-removed (proxy for how much draft the human
reads at each stage; grows linearly with step index $t$). The dashed Pareto
hull connects \textsc{Burst-Early} and \textsc{Spread}, the two oversight schedules
retained by lower convex-envelope extraction; \textsc{Skip} (cost~$0$) is shown for
reference but excluded from the hull and the $\lambda$ strip. The bottom
strip shows which schedule minimizes $\mathcal{L}_\lambda(S) = (10-\text{quality})
+ \lambda \cdot \text{cost}$ as the cost weight $\lambda$ increases. \textsc{Spread} wins up to $\lambda \approx 3.3\times10^{-4}$,
beyond which the schedule \textsc{Burst-Early} wins.}
\label{fig:study1}
\end{figure}

\textbf{Results.}
Table~\ref{tab:results-study1} reports mean quality and cost for each schedule. Figure~\ref{fig:study1} plots all six schedules in quality-cost space with the
Pareto hull and the optimal schedule under different values of $\lambda$. Figure~\ref{fig:study1comp} gives a qualitative comparison for one example paper under four distinct oversight schedules.
 
\begin{table}[htbp]
\centering
\small
\caption{Study~1 results: writing related work with $N=40$ ICLR~2026 papers. Cost is the mean paper-level oversight cost, measured by the total length of draft content read at the schedule's oversight stages.
Quality is the mean paper-level score, where each paper-level score averages the judge scores for coverage and factual accuracy. (1--10
scale).
A schedule is on the Pareto hull if it minimizes the loss
${\mathcal{L}}_\lambda(S) = (10-\text{quality}) + \lambda\cdot\text{cost}$ for some $\lambda \ge 0$ among the five schedules with $K=3$.}
\begin{tabular}{lccc}
\toprule
\textbf{Schedule} & \textbf{Cost (mean)} & \textbf{Quality (mean $\pm$ SE)} & \textbf{On Pareto hull?} \\
\midrule
\textsc{Skip}                        & $0$    & $3.93 \pm 0.16$ & --- \\
\textsc{Burst-Early}                 & $511$  & $4.86 \pm 0.18$ & \ding{51} \\
\textsc{Tilt-Early}                  & $819$  & $4.94 \pm 0.12$ & \ding{55}  \\
\textsc{Spread}            & $1122$ & $5.06 \pm 0.18$ & \ding{51} \\
\textsc{Uniform}                     & $1180$ & $5.01 \pm 0.13$ & \ding{55}  \\
\textsc{Burst-Late}                  & $1797$ & $5.05 \pm 0.12$ & \ding{55}  \\
\bottomrule
\end{tabular}

\label{tab:results-study1}
\end{table}

The Pareto hull of study~1 comprises $\{\textsc{Burst-Early},\;\textsc{Spread}\}$.
\textsc{Spread} achieves the best quality score $5.06$ in this study at the cost of $1122$ tokens. These optimal schedules with non-decreasing gaps inspired the nonuniformity principle.

%-----------------------------------------------------------------------
\subsection{Study 2: constructing an HTML page}
\label{subsec:study2}

This study focuses on human-AI coworking for constructing an HTML page.
We present the setup, metrics, and results below, and the full implementation details of this study are in Section~C of the supplementary material.

\textbf{Setup.}
We construct $N=10$ tasks, each consisting of an initial prompt and a design intent document. For each task instance, we set up three roles, agent for building the page (writing the HTML code), human for providing oversight, and judge for giving quality scores on the final pages, each realized with separate LLM or vision language model calls  with temperature $=0$:
\begin{itemize}
\item \textbf{Agent} (building the HTML page, \texttt{deepseek-v4-flash}, text-only): given only the initial prompt, constructs the landing page section by section over $T=10$ build steps, where each step adds one major section. At any oversight stage, the agent revises the page based on human oversight.
\item \textbf{Human} (providing oversight, \texttt{claude-sonnet-4-6}, vision): at any oversight stage $s\in S$, given the design intent document as specification $\theta$ and the rendered screenshot of the HTML code produced up to stage $s$, returns feedback to the agent with suggestions for revision and guidance on constructing the remaining parts.
\item \textbf{Judge} (giving quality score, \texttt{claude-sonnet-4-6}, vision): given the final rendered pages under all six schedules, scores (1--10 integer scale) each page on intent alignment and visual hierarchy.
\end{itemize}

For each schedule listed in Table~\ref{tab:schedules-all}, we measure quality through the judge, and cost through the depth of the page build at which human oversight occurs, as explained in detail below.

\textbf{Quality.}
Two dimensions are scored by the judge:\textbf{hidden intent alignment} (does the page match the tone, required emphasis, constraints, and section ordering in the design intent document?) and \textbf{visual hierarchy} (is the layout clear, well-spaced, scannable, and effective in emphasizing calls to action?). After all final pages are produced, for each task, the judge scores the pages produced under the six schedules. To reduce the effect of presentation order, we repeat this scoring three times, each time with a randomized order of the six pages presented to the judge. Then the medians across the three repetitions are set as the quality score for each dimension. The task-level quality per schedule is then given by the average of the two dimensions' quality scores.

\textbf{Cost.}
Human oversight cost is directly set to $s$ at an oversight stage $s$. This reflects the amount of accumulated work that the human must inspect when the review occurs. The cost for an oversight schedule $S$ is then given by $\widehat{C}(S)=\sum^K_{k=1}s_k$.
This is a fixed schedule-level proxy.

\begin{figure}[htbp]
\centering
\includegraphics[width=1\textwidth]{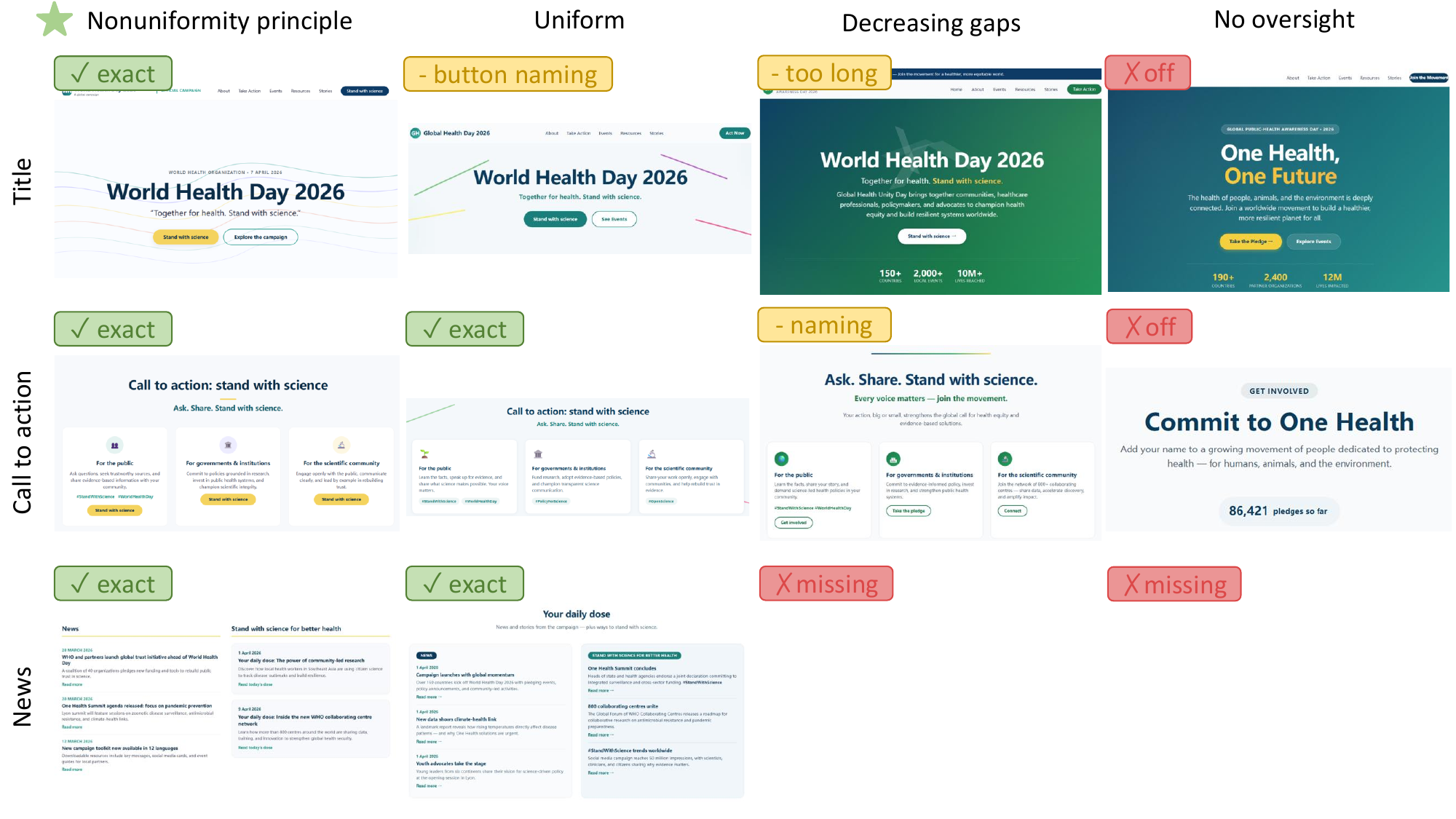}
\caption[]{Qualitative comparison of one example from Study~2, showing the final HTML pages produced under four distinct schedules.
Each column corresponds to one schedule, and each row compares one part of the page: title, call to action, and news section.
The schedule satisfying the nonuniformity principle (\textsc{Tilt-Early}, with increasing gaps) performs best: it uses the correct World Health Day 2026 title, preserves the required call to action framing, and includes the required news section.
The schedule with uniform gaps (\textsc{Uniform}) is mostly correct, but it makes a small error through incorrect button naming at the upper right corner.
The schedule with decreasing gaps (\textsc{Burst-Late}) is less well aligned: its title region is too long, its call to action uses incorrect naming, and the news section is missing.
The schedule with no oversight (\textsc{Skip}) performs worst: it has a wrong campaign title, a wrong call to action section, and a missing news section.}
\label{fig:study2comp}
\end{figure}

\begin{figure}[htbp]
\centering
\includegraphics[width=0.85\textwidth]{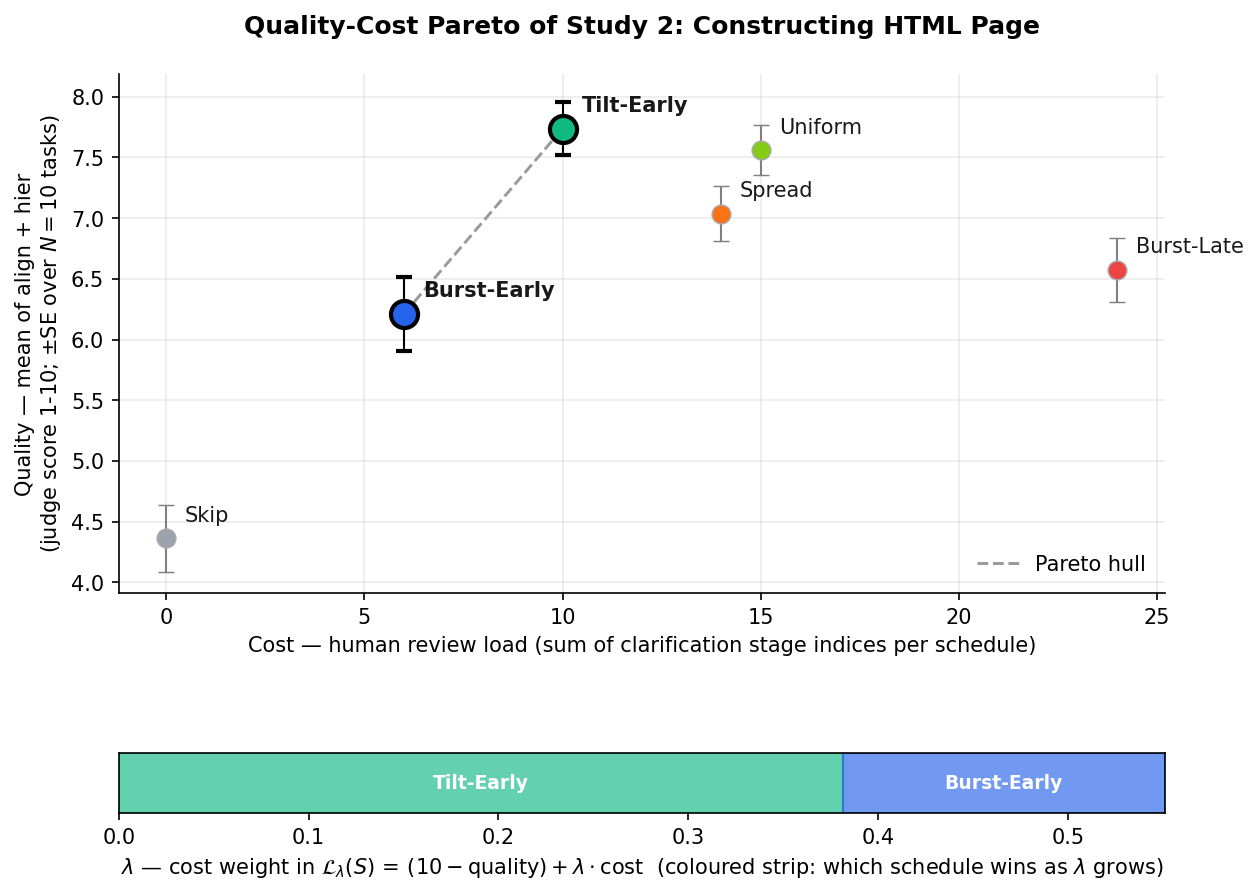}
\caption[]{Quality-cost Pareto frontier of study~2: HTML landing-page construction over
$N=10$ tasks. Each point is one oversight schedule; error bars show $\pm 1$~SE
across tasks ($\mathrm{SE} = s/\sqrt{10}$).\textbf{Quality} (vertical axis): mean
of median hidden-intent alignment and visual-hierarchy judge scores (1--10 scale).\textbf{Cost} (horizontal axis): human review load $\sum t$, the sum of oversight
step indices (a fixed schedule constant proxying how deep into the build the reviewer
must engage). The dashed Pareto hull connects \textsc{Burst-Early} and \textsc{Tilt-Early},
the two oversight schedules retained by lower convex-envelope extraction; \textsc{Skip}
(cost~$0$) is shown for reference but excluded from the hull and the $\lambda$ strip.
The bottom strip shows which schedule minimises
$\mathcal{L}_\lambda(S) = (10-\text{quality}) + \lambda \cdot \text{cost}$ as the
cost weight $\lambda$ increases. \textsc{Tilt-Early} wins up to $\lambda \approx 0.38$,
beyond which the schedule \textsc{Burst-Early} wins.}
\label{fig:study2}
\end{figure}

\textbf{Results.}
Table~\ref{tab:results-study2} reports mean quality and cost for each schedule. Figure~\ref{fig:study2} plots all six schedules in quality-cost space with the Pareto hull and the optimal schedule under different values of $\lambda$. Figure~\ref{fig:study2comp} gives a qualitative comparison for one example HTML task under four distinct oversight schedules.

The Pareto hull comprises $\{\textsc{Burst-Early},\;\textsc{Tilt-Early}\}$: \textsc{Burst-Early} has a lowest cost, $6$ and \textsc{Tilt-Early} has the highest quality score of $7.74$. \textsc{Skip}
is plotted as a reference point but excluded from the hull, which is computed over the five
oversight schedules.
\textsc{Tilt-Early} achieves the highest quality among all oversight schedules ($7.74\pm
0.21$) at moderate cost ($\sum t = 10$). \textsc{Uniform} spends 50\% more in oversight cost
($\sum t=15$) for lower quality ($7.56$). \textsc{Burst-Late} delivers the worst quality
among oversight schedules ($6.58$) at the highest cost ($\sum t=24$), despite receiving
the same number of human calls ($K=3$).

\begin{table}[htbp]
\centering
\small
\caption{Study~2 results: HTML landing-page construction over $N=10$ tasks. Quality is the
mean page-level score, where each page-level score averages the judge scores for hidden intent alignment and visual hierarchy (1--10 scale).
Bold denotes the highest quality scores. A schedule is on the Pareto hull if it minimizes the loss
${\mathcal{L}}_\lambda(S) = (10-\text{quality}) + \lambda\cdot\text{cost}$ for some $\lambda \ge 0$ among the five schedules with $K=3$.}
\begin{tabular}{lccc}
\toprule
\textbf{Schedule} & \textbf{Cost $(\sum t)$} & \textbf{Quality (mean $\pm$ SE)} & \textbf{On Pareto hull?} \\
\midrule
\textsc{Skip}       & $0$  & $4.36\pm 0.28$ & --- \\
\textsc{Burst-Early} & $6$  & $6.21\pm 0.30$ & \ding{51} \\
\textsc{Tilt-Early} & $10$ & $\mathbf{7.74\pm 0.21}$ & \ding{51}  \\
\textsc{Spread}     & $14$ & $7.04\pm 0.23$ & \ding{55}   \\
\textsc{Uniform}    & $15$ & $\mathbf{7.56\pm 0.20}$ & \ding{55}  \\
\textsc{Burst-Late}  & $24$ & $6.58\pm 0.26$ & \ding{55}   \\
\bottomrule
\end{tabular}
\label{tab:results-study2}
\end{table}

%-----------------------------------------------------------------------
\subsection{Discussion of experimental results in relation to the theory}\label{subsec:discussion}
For the experimental results shown in Figure~\ref{fig:study1} and
Figure~\ref{fig:study2}, we
validate them with the theoretical results in Section~\ref{sec:oversight-principle}. The three empirically optimal schedules have
$S_{\mathrm{spread}}=(1,4,9),$
$\bm d_{\mathrm{spread}}=(1,3,5,1),$
$S_{\mathrm{tilt}}=(1,3,6),$
$\bm d_{\mathrm{tilt}}=(1,2,3,4),$ and $S_{\mathrm{early}}=(1,2,3),$
$\bm d_{\mathrm{early}}=(1,1,1,7),$ respectively.
All of them satisfy
$d_0\le d_1\le d_2,$
which fits the main conclusion of Theorem~\ref{thm:nonuniformity}. Moreover, in study~1 when $\lambda>3.4\times10^{-4}$, and in study~2 when $\lambda > 0.39$, the schedule \textsc{burst-early} dominates. This aligns with Corollary~\ref{cor:all-early-linear-cost} from Theorem~\ref{thm:all-early-general-cost}.

We also validate Algorithm~\ref{alg:rw-schedule} and Proposition~\ref{prop:rw-schedule-algorithm} by deriving that under certain values of $\kappa$ and $\eta$, Algorithm~\ref{alg:rw-schedule} will return $S_{\mathrm{spread}}=(1,4,9)$ or $S_{\mathrm{tilt}}=(1,3,6)$ as a minimizer. This result is given as follows, and the derivation details are given in Section~D of the supplementary material.
For $T=10$ and $K=3$, Algorithm~\ref{alg:rw-schedule} returns
$S_{\mathrm{spread}}=(1,4,9)$ as a minimizer whenever
\begin{equation}
\label{eq:spread-region}
\max\left\{
1-6\kappa,\,
\frac{1-4\kappa}{2},\,
\frac{1-2\kappa}{3},\,
\frac{3\kappa}{2}
\right\}
\le
\eta
\le
\min\left\{
2-5\kappa,\,
\frac{2-3\kappa}{2},\,
3\kappa
\right\}.
\end{equation}
The region~\eqref{eq:spread-region} is nonempty for
$\frac{1}{9}\le \kappa\le \frac{4}{13}.$
For example, $\kappa=0.2$ and $\eta=0.4$ satisfy
\eqref{eq:spread-region}, and Algorithm~\ref{alg:rw-schedule} returns
$\widehat{\bm d}=(1,3,5,1)$ and 
$\widehat S=(1,4,9).$

Similarly, Algorithm~\ref{alg:rw-schedule} returns
$\bm d_{\mathrm{tilt}}=(1,2,3,4)$ as a minimizer whenever
\begin{equation}
\label{eq:tilt-region}
\max\left\{
4-4\kappa,\,
\frac{4-3\kappa}{2},\,
\frac{4-2\kappa}{3},\,
\frac{\kappa}{2}
\right\}
\le
\eta
\le
\min\left\{
2\kappa,\,
5-3\kappa,\,
\frac{5-2\kappa}{2}
\right\}.
\end{equation}
The region~\ref{eq:tilt-region} is nonempty for
$\frac{2}{3}\le \kappa<1.$
For example, $\kappa=0.7$ and $\eta=1.3$ satisfy
\eqref{eq:tilt-region}, and Algorithm~\ref{alg:rw-schedule} returns
$\widehat{\bm d}=(1,2,3,4),
\widehat S=(1,3,6).$ Finally, the burst-early
schedule with
$S_{\mathrm{early}}=(1,2,3)$ and
$\bm d_{\mathrm{early}}=(1,1,1,7)$ is a minimizer returned by Algorithm~\ref{alg:rw-schedule} whenever
$\eta>7-2\kappa.$ This also agrees qualitatively with the empirical frontier:
as $\lambda>3.4\times10^{-4}$ in study~1 and $\lambda>0.39$ in study~2, the optimal schedule is \textsc{burst-early}.

Overall, the experimental results align with the nonuniformity principle. Across the two studies, the optimal schedules have non-decreasing gaps, which agrees with the main result of Theorem~\ref{thm:nonuniformity}. The validation based on Algorithm~\ref{alg:rw-schedule} further shows that the empirically favorable schedules also arise from the proposed practical guide under appropriate parameter regions.

\section{Conclusion}
\label{sec:conclusion}
% We propose the \emph{nonuniformity principle} for scheduling human oversight in human-AI coworking. The principle states that, when misalignment accumulates as AI works and reviewing a later deliverable is more costly to the human, human oversight should be concentrated earlier and become progressively less frequent as the task proceeds. We formalize human-AI coworking as a process where AI constructs a deliverable and receives oversight from the human at a fixed number of selected stages. The objective is then to find the best schedule that minimizes a combination of the alignment quality between the final deliverable and the human's requirement, and the oversight cost. We then derive the nonuniformity principle, which states that the optimal oversight schedule has non-decreasing gaps. The principle is motivated and supported by two empirical studies, where schedules with non-decreasing gaps achieve favorable quality-cost trade-offs.

We formalize human-AI coworking as a process where an AI constructs a deliverable and receives oversight from the human at a given number of selected stages. The objective is to find the best schedule that minimizes a combination of the alignment loss between the final deliverable and the human's requirement, and the oversight cost. We propose the \emph{nonuniformity principle}: when misalignment accumulates as the AI works and reviewing a later deliverable is more costly, the optimal oversight schedule has non-decreasing gaps. That is, human oversight is better concentrated earlier and become progressively less frequent as the task proceeds. The principle is motivated and supported by empirical studies, where schedules with non-decreasing gaps achieve favorable quality-cost trade-offs.

We suggest two directions for future research. First, this paper studies the timing of human oversight. A broader framework could jointly determine {when} oversight should occur, {what} aspects of the intermediate deliverable should be reviewed, and {how} feedback should be provided. Such a framework could also allow the number of oversight stages itself to be adaptive.
Second, AI-for-science workflows such as hypothesis generation and experimental design present a high-stakes setting where the nonuniformity principle applies
directly: these tasks are long-horizon, specification-rich, and expensive to review in full. Empirical studies in scientific discovery pipelines would test the principle's scope and inform the design of oversight-aware autonomous research agents.

\section*{Use of Generative AI Tools}

During the preparation of this manuscript, the authors used AgentLab (MorphMind) to prototype the oversight scheduling pipelines in the experiments, design test cases for the algorithm, and assist with figure design. Claude Opus 4.8 (Anthropic) was used for coding assistance and language improvement. 
The authors reviewed and edited all outputs and take full responsibility for the content of this manuscript.

%%%%%%%%%%%%%%%%%%%%%%%%%%%%%%%%%%%%%%%%%%%%%%%%%%%%%%%%%%%%%%%%%%%%%%%%%%%%%%
\bibliographystyle{plainnat}
\bibliography{refs} % Please add your references in this BibTeX file with appropriate placeholders [REF]

\end{document}